\documentclass{article} 
\usepackage{iclr2027_conference,times}

\usepackage{amsmath,amsfonts,bm}

\def\eqref#1{equation~\ref{#1}}

\def\1{\bm{1}}

\DeclareMathAlphabet{\mathsfit}{\encodingdefault}{\sfdefault}{m}{sl}
\SetMathAlphabet{\mathsfit}{bold}{\encodingdefault}{\sfdefault}{bx}{n}

\usepackage{hyperref}
\usepackage{url}
\usepackage{graphicx}
\usepackage{booktabs}
\usepackage{multirow}
\usepackage{amssymb}
\usepackage{xcolor}
\usepackage{colortbl}
\usepackage{wrapfig}
\usepackage{flafter}
\usepackage[most]{tcolorbox}
\definecolor{oursblue}{RGB}{219,238,247}
\newcommand{\venue}[1]{{\fontsize{4}{5}\selectfont\textcolor{gray}{[#1]}}}

\tcbset{
  promptbox/.style={
    enhanced, breakable, colback=gray!5, colframe=black!70,
    boxrule=0.6pt, arc=2pt, left=6pt, right=6pt, top=4pt, bottom=4pt,
    fonttitle=\bfseries\small, coltitle=white, colbacktitle=black!70,
    attach boxed title to top left={xshift=4pt, yshift=-2pt},
    boxed title style={colback=black!70, sharp corners},
    fontupper=\scriptsize\ttfamily,
  }
}

\title{EGSD: Event-Grounded Self-Distillation for Streaming Video Understanding}

\author{
\textbf{Yuwei Miao}$^{1,*}$, \textbf{Xuesheng Zhang}$^{1,*}$, \textbf{Wenhao Zou}$^{2}$, \textbf{Jixia Zhang}$^{3}$, \textbf{Jianwei Lv}$^{1}$, \\
\textbf{Bo Yuan}$^{1}$, \textbf{Junfeng Wang}$^{1}$, \textbf{Shiao Xie}$^{1,\dagger}$ \\
$^{1}$Baidu \quad $^{2}$University of Chinese Academy of Sciences \quad $^{3}$China University of Petroleum, Beijing
}

\newcommand\blfootnote[1]{%
  \begingroup
  \renewcommand\thefootnote{}%
  \footnotetext{#1}%
  \addtocounter{footnote}{-1}%
  \endgroup
}

\iclrfinalcopy 
\begin{document}

\maketitle
\lhead{Preprint}

\blfootnote{$^{*}$Equal contribution \quad $^{\dagger}$Corresponding author}

\begin{abstract}
Real-time video understanding requires incrementally maintaining a memory of
streaming content, and optimizing this requires
dense process signals. On-Policy Self-Distillation (OPSD), which lets one model
serve as both teacher and student with the teacher receiving additional privileged
information such as the question and ground-truth (GT) answer, can supply such
token-level signals.
However, applying it directly to streaming video raises two problems.
(1) The student cannot be optimized end-to-end, where memory is written before the
question arrives, yet the teacher scores it with the question-and-GT privilege,
misaligning their preferences.
(2) Effective-entity memory collapses, where the question-and-GT
privilege makes the teacher favor only question-relevant entities, and token-mean
averaging over a memory renders its signal invariant to how many entities that memory
covers, both driving memory against the streaming need for diversity.
To address these issues, we propose
\textbf{Event-Grounded Self-Distillation (EGSD)}, which characterizes streaming
memory as an incremental update over verifiable Events (key visual entities,
actions, and details) and targets the two problems on this basis.
For problem (1), we adapt the OPSD signal into a multiplicative weight combined
with the outcome reward; for problem (2), we
re-weight the teacher with Events as privileged information to counter its
question-relevance bias, and add an entity-coverage reward to supply the coverage
preference the token-mean teacher lacks. Extensive experiments on mainstream online
and offline benchmarks show EGSD achieves strong performance, reaching 79.8\% on
StreamingBench and 73.4\% on the OVO-Bench Real-Time track, while memory analysis
shows effective-entity recall rises 17.4\% at only 6.8\% more memory length.
\end{abstract}

\section{Introduction}
\label{sec:intro}

Video Large Language Models (VideoLLMs) have achieved significant success in
offline video understanding
\citep{chen2025longvila,wang2025internvideo2,qin2025videoxl2longvideounderstandingtaskaware}.
However, real-world applications such as live-stream QA and embodied
intelligence \citep{chen2024videollm,chen2025live,driess2023palm} demand real-time
processing of streaming input, bound by temporal causality and a limited context
window. This raises a central challenge, incrementally memorizing long-horizon
visual context while producing causal responses in an online manner
\citep{lin2026streamingbench}.

Such incremental memory of the streaming history admits different representations. One way compresses visual
information at the token or KV-cache level or into hidden states, and
is mainly optimized by training-free or lightweight-adapter
inference that prunes visual tokens by inter-frame similarity, attention
importance, and similar criteria
\citep{kim2026infinipot,xie2026fluxmem,xu2026streamingvlm}; yet
such memory is not human-readable, so whether it retains the information a future
query needs cannot be measured explicitly. The other way writes the history as
natural-language text, which is reviewable and is the memory this
work adopts. Existing methods optimize such memory with an SFT-RL pipeline
\citep{guan2026video,xie2026streamrag}, but its reward
is verifiable only at
the end of a long trajectory, leaving the model passively driven by final-answer
correctness.

On-Policy Distillation (OPD) \citep{gu2023minillm,agarwal2024policy} is an
important post-training paradigm that supplies dense, token-level process signals
within a trajectory. Vanilla OPD needs a larger same-family teacher, bounding
gains by the teacher's capability at high cost; On-Policy Self-Distillation (OPSD)
\citep{zhao2026self} instead lets a student holding privileged information act as
its own teacher, needing no larger model and balancing training cost against
performance. Recent work explores this paradigm across image-text and
mathematical reasoning with RLSD \citep{yang2026self}, multi-turn agents
with StepOPSD \citep{zhang2026stepopsd}, multimodal reasoning with OPLD
\citep{zhu2026opld}, fine-grained regional perception with Vision-OPD
\citep{yuan2026vision}, and privileged visual evidence with Video-OPSD
\citep{wang2026video}, differing in what privileged information
the teacher sees and how the resulting signal re-weights tokens; among these, the
mainstream privileged information is the final question and the ground-truth
answer.

However, these OPSD methods mostly target single-turn or offline settings, whereas
streaming video understanding with natural-language memory has two distinctive
traits. First, the intermediate memory turns run before any question, so the model
can only record the salient visual content it currently observes. Second, for
low-latency answering the model must reply from its existing textual memory and the
current video clip alone, so the diversity of effective visual entities in memory is
critical. Directly applying existing OPSD methods to streaming video thus raises
two problems.
\begin{wrapfigure}{r}{0.46\textwidth}
\vspace{-0.9\baselineskip}
\centering
\includegraphics[width=\linewidth]{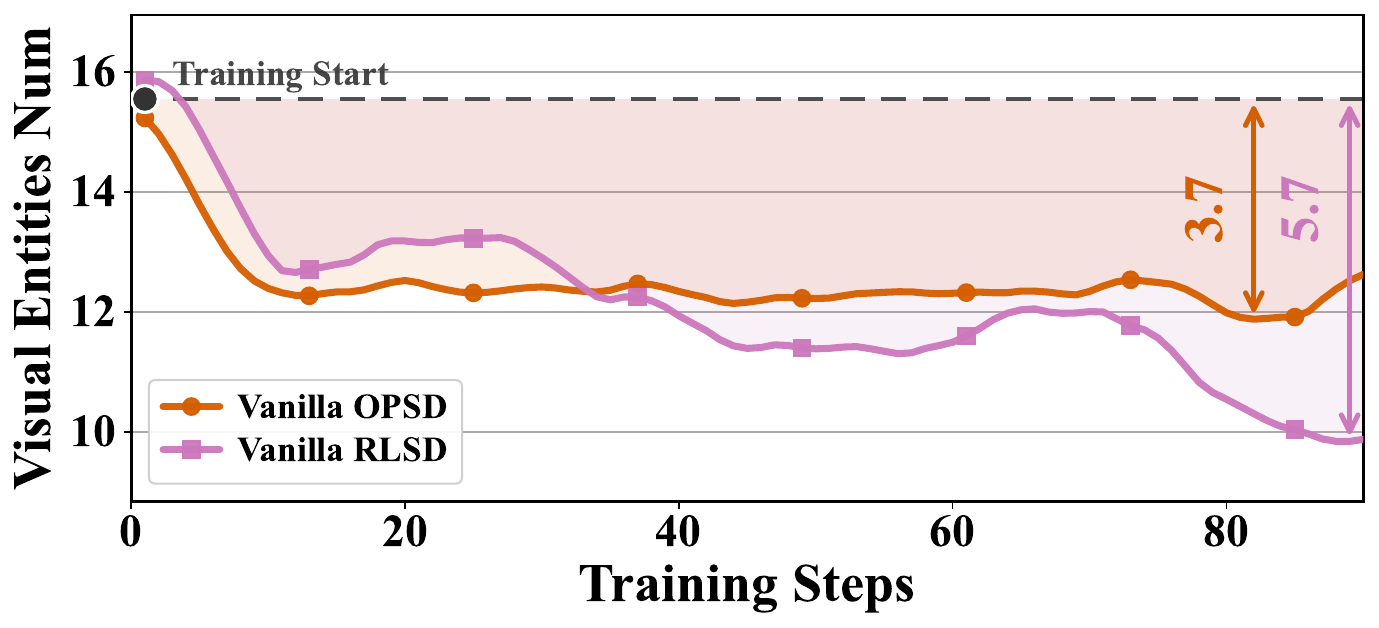}
\caption{Applying OPSD/RLSD to streaming video collapses memory diversity. }
\label{fig:intro_collapse}
\vspace{-0.7\baselineskip}
\end{wrapfigure}
(1) \textbf{The student cannot be optimized end-to-end.} In single-turn QA the teacher
adds the ground-truth answer to the shared question, so distilling it aligns with
answering correctly; in streaming, memory is written before the question arrives, so
the student writes from its own visible state while the teacher scores with the
question known, misaligning the teacher's preference with optimizing memory for final performance.
(2) \textbf{Effective-entity memory collapses.} As shown in Figure~\ref{fig:intro_collapse},
the student memory retains far fewer visual entities as training proceeds, driven by
two causes. First, the existing question-and-GT
privilege makes the teacher favor question-useful entities and discriminate against
equally real but irrelevant ones. Second, an
intrinsic OPSD limitation is that, under token-mean averaging over a memory, the
teacher signal is invariant to how many effective entities a rollout grounds, so it
cannot prefer a higher-coverage memory over one grounding fewer entities. Both push
memory opposite to the streaming goal of retaining as much key visual content as
possible.

To address these problems, we propose Event-Grounded Self-Distillation (EGSD).
Inspired by the event-segmentation theory in cognitive science
\citep{zacks2001event,zacks2007event}, we argue that streaming video
understanding based on incremental textual memory is essentially an incremental
update over Events, comprising three complementary parts: key visual entities,
actions, and details (e.g., spatial relations and OCR). Unlike methods that
finely optimize the memory format~\citep{song2024moviechat,diko2025rewind}, an Event is a
deterministic signal that can be extracted offline, verified, and defined
independently of any question, giving a stable grounding basis for
the teacher and reward. EGSD strengthens the diversity of streaming memory
and the final answering accuracy end-to-end by targeting the two problems. For problem (1) we introduce
\textbf{Multiplicative teacher gating}, where we analyze why the OPSD signal fails in
the streaming-video setting and establish that combining it as a multiplicative weight
with the outcome reward is the right form for end-to-end streaming
optimization that keeps the student learnable from its own memory. For problem (2) we introduce
\textbf{Event-privileged teacher re-weighting}, where on top of the question and GT
answer we add a verifiable set of Event facts to the teacher's privileged
information, preventing the teacher's reward from attending only to
question-relevant entities and thereby avoiding memory collapse, together with an
\textbf{Entity-coverage reward} that explicitly prefers rollouts covering more of the clip's
effective entities, supplying the coverage preference the token-mean teacher lacks.

EGSD's mechanisms are complementary in streaming video: the Event-based reward drives the model
to write more effective entities, while multiplicative gating keeps this optimization aligned with
the final answer, so their interaction lifts performance. Our contributions are as follows:

\begin{itemize}
\item We analyze and verify the causes of end-to-end optimization failure and
memory collapse when vanilla OPSD is applied directly to streaming video understanding.
\item We propose EGSD, which identifies verifiable Events as the essential content
of streaming memory and strengthens memory diversity and answering accuracy
end-to-end through multiplicative gating, Event-privileged re-weighting, and an
entity-coverage reward.
\item EGSD achieves strong performance on online and offline benchmarks, and memory
analysis attributes this to entity recall rising $17.4\%$ at only $6.8\%$ more memory length.

\end{itemize}

\section{Related Work}

\subsection{Multimodal LLMs for Long Video and Streaming Video Reasoning}

Offline long-video LLMs \citep{yue2025mimo,li2024monkey} scale along context
extension \citep{chen2025longvila} and token compression
\citep{wang2025internvideo2,qin2025videoxl2longvideounderstandingtaskaware}, but
assume the full video is available, whereas streaming exposes neither future
frames nor query time. Among causal single-pass methods for streaming video, the history representation
determines whether an external function can inspect it.
Visual-token or KV compression
\citep{yao2025timechat,kim2026infinipot,xie2026fluxmem,xu2026streamingvlm} drops
content irrecoverably; hidden-state memory
\citep{chen2024videollm,qian2025dispider,zeng2026streamforest,wang2025videollamb,sun2025video}
is compact but unreadable and unscorable; natural-language memory
\citep{guan2026video,xie2026streamrag} is readable and reusable at a token cost.
However, none of these imposes a trainable, reward-checkable objective on the memory
content itself, which is the premise of our Event-grounded memory.

\subsection{Video Reinforcement Learning}

Recent methods for RL in long-video reasoning augment the outcome reward with an
additional process reward, since over a long horizon the final answer alone cannot
tell whether a failure comes from looking at the wrong moment or from flawed reasoning. Yet these rewards score
``where to look'' rather than ``what to remember'', whether via key-frame
grounding \citep{cao2025videominer}, zoom-in \citep{fu2025love}, temporal
grounding \citep{liu2026videotemp}, or sampler-MLLM joint optimization
\citep{tan2026msjoe}, all evaluating a coordinate against interval annotations. In
contrast, none rewards the content of streaming memory, which is exactly where our
process reward applies.

\subsection{Multimodal On-Policy Self-Distillation}

OPSD lets a privileged copy of the student act as its own teacher, giving dense
supervision without a larger model. Across all existing instantiations, and this
is what matters for streaming, the tokens the teacher scores are the same tokens
that produce the answer being evaluated, whether the target is a
single-turn image \citep{zhu2026opld}, a reasoning chain \citep{yang2026self}, an
agent rollout \citep{zhang2026stepopsd}, a crop-conditioned fine-grained answer
\citep{yuan2026vision}, or an evidence-grounded video answer
\citep{wang2026video}, so the teacher judges tokens the student already produced
toward a question it already knows. Streaming memory writing breaks this premise,
as memory turns are written before any question arrives and useful content is
defined by future queries, so no existing OPSD signal can supervise it, the gap
this work closes.

\section{Method}

\begin{figure}[t]
\begin{center}
\includegraphics[width=\textwidth]{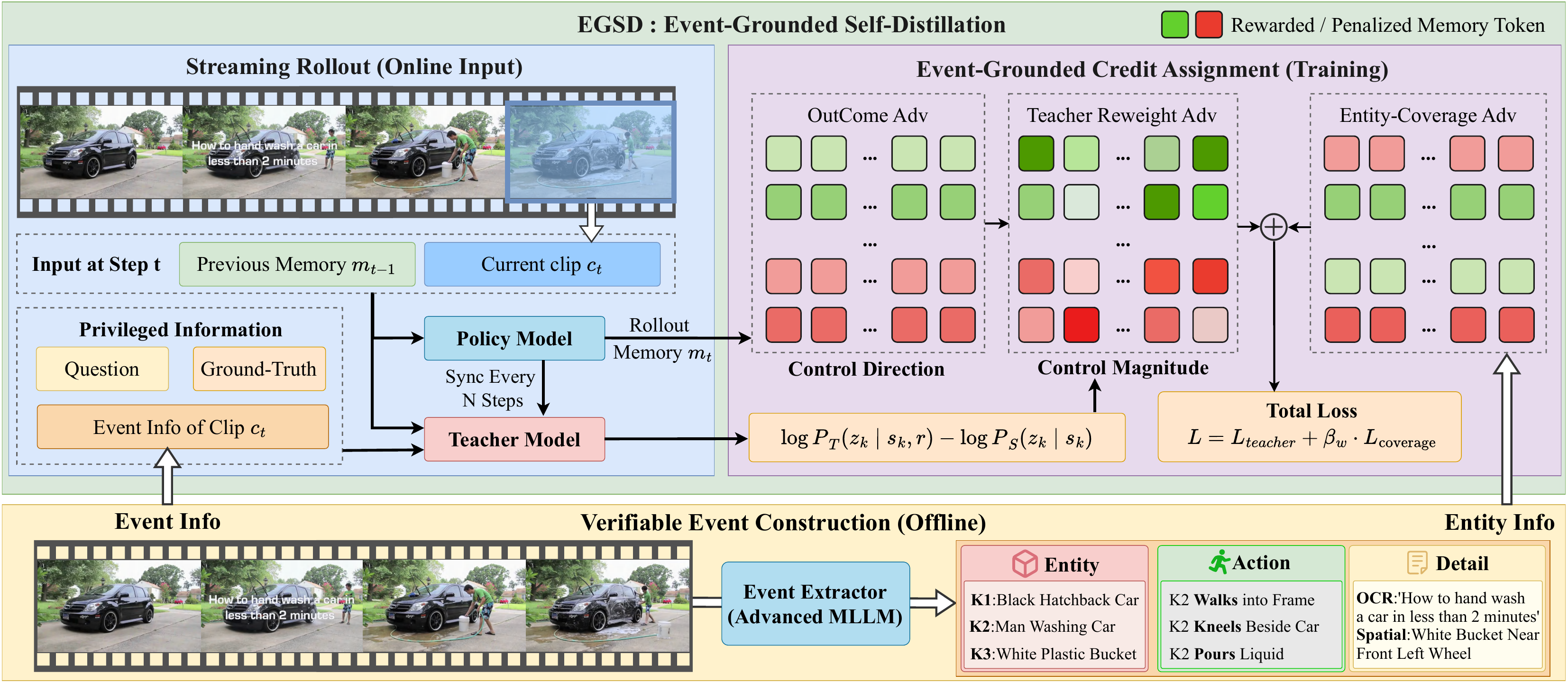}
\end{center}
\caption{Overview of EGSD. The student writes free-text memory per clip without
seeing the question, while a frozen large model extracts a per-clip Event fact set
offline. The Event-privileged teacher re-weights student tokens into a
multiplicative advantage, and an Event-based entity-coverage reward prefers
higher-coverage memories, guiding memory toward grounded diversity.}
\label{fig:framework}
\end{figure}

\subsection{Analysis of the OPSD Signal in the Streaming-Video Setting}

\paragraph{Streaming protocol.} We adopt a
multi-turn streaming protocol that splits the video stream into $K$
sequentially arriving clips by cumulative visual-token capacity. For the first
$K-1$ clips, without knowing the final question $q$, the model generates a
streaming thought $z_k$ from the current visual content $c_k$ and the historical
memory $m_{k-1}$ and updates the memory, and when the $K$-th clip arrives and $q$
becomes visible it produces the final answer $a$ from the accumulated memory,
\begin{equation}z_k\sim\pi_\theta(\cdot\mid c_k,m_{k-1}),\qquad m_k=\mathrm{Update}(m_{k-1},z_k).
\end{equation}

\paragraph{OPSD as a teacher-provided process signal.} Vanilla OPD
supplies a per-token process signal through the reverse KL between a student $P_S$
and a same-family teacher $P_T$ conditioned on the same prior trajectory,
\begin{equation}\mathrm{KL}(P_S\,\|\,P_T)=
\mathbb{E}_{P_S}\!\left[\log P_S(y_t\mid y_{<t})
-\log P_T(y_t\mid y_{<t})\right].
\end{equation}OPSD instantiates the teacher as the student itself additionally conditioned on
privileged information $\psi$, i.e., $P_S(z_k)=\pi_\theta(z_k\mid s_k)$ and
$P_T(z_k)=\pi_\theta(z_k\mid s_k,\psi)$ with $s_k=(c_k,m_{k-1})$ the state visible to
the student at clip $k$, so the per-memory signal becomes
\begin{equation}\delta_k=\log P_T(z_k\mid s_k,\psi)-\log P_S(z_k\mid s_k),
\end{equation}which measures the log-probability gain the privileged information brings to this
memory.

\paragraph{Multiplicative OPSD fits streaming video.} The distinctive challenge of
streaming is that the student writes memory before the question arrives, whereas
the teacher scores it with the question already known, so $\delta_k$ prefers an
answer-aware memory that is misaligned with the student's and cannot drive
end-to-end optimization on its own. The difficulty is thus how to
propagate the outcome to early memory decisions while preserving the teacher's
clip-level resolution. Additive schemes such as VOLD \citep{bousselham2026vold},
which sum a distillation loss with a GRPO objective, fall short here, as the loss
ratio is hard to tune into cooperation toward end-to-end optimization.

We thus propose the key viewpoint that credit assignment in streaming video should
let the final result decide the update direction and the privileged teacher the
credit magnitude across memory decisions. Accordingly, we adopt outcome-conditioned
multiplicative teacher modulation,
\begin{equation}\hat A_k=A_R\cdot\rho_k,\qquad \rho_k>0,
\end{equation}where the positivity of $\rho_k$ keeps $\hat A_k$ aligned in sign with $A_R$,
so $A_R$ alone sets the update direction while the teacher gain $\delta_k$, through
$\rho_k$, only modulates the credit magnitude across memory decisions.

\subsection{EGSD: Event-Privileged Teacher Re-weighting and Entity-Coverage Reward}

\S3.1 established multiplicative teacher modulation as the right form of credit
assignment for streaming memory but left open what the teacher should be
conditioned on. As analyzed in \S1, mainstream OPSD conditions the teacher on the
question and GT answer, rewarding only question-relevant memory and progressively
collapsing the student's output space, at odds with the streaming need for diverse memory.
We therefore add a verifiable set of objective facts, termed the Event, to the
teacher's privilege, extending its discriminative axis from question relevance to
faithfulness to what has happened. As shown in Figure~\ref{fig:framework}, \S3.2.1 introduces the Event representation,
\S3.2.2 derives a privileged teacher anchored on fact grounding, and \S3.2.3
develops a coverage reward that ranks memories by how many effective entities they
cover.

\subsubsection{Event: A Triple-Component Verifiable Signal}

Free-text memory $z_k$ is a fluent carrier of a video clip, yet a plain narrative
carries no structure verifiable against the visual content, so both the teacher and
the reward are left to discriminate on an unverifiable textual surface. We therefore decompose the Event of the $k$-th clip into three complementary
components, extracted offline by a frozen large model and collected into a per-clip
fact set,
\begin{equation}\mathcal{F}_k=\mathcal{E}_k\cup\mathcal{A}_k\cup\mathcal{D}_k,
\end{equation}where $\mathcal{E}_k$ (\texttt{ENTITY}) are key visual entities such as people,
things, and text signs, $\mathcal{A}_k$ (\texttt{ACTION}) are temporally ordered
behaviors among them, and $\mathcal{D}_k$ (\texttt{DETAIL}) are verbatim-checkable
attributes such as OCR text, counts, and spatial relations. This
decomposition breaks a single free-text narrative into typed fact units that can be
verified independently, and we use $\mathcal{F}_k$ as the privileged information on
which the teacher of \S3.2.2 and the coverage reward of \S3.2.3 give cleaner, more
grounded signals.

\subsubsection{Event-Based Privileged Teacher}

\paragraph{Adding Events to the privileged context.} We augment the teacher's
privileged context with the per-clip fact set
$\mathcal{F}_k$ from \S3.2.1, so the teacher and student conditions are
\begin{equation}P_T(\cdot\mid y_{<t}) \triangleq \pi_\theta(\cdot\mid s_k, \mathcal{F}_k, q, a^\star, y_{<t}),
\qquad
P_S(\cdot\mid y_{<t}) \triangleq \pi_\theta(\cdot\mid s_k, y_{<t}),
\end{equation}which share the same student state $s_k$ and differ only in whether the teacher sees
$(\mathcal{F}_k,q,a^\star)$. Adding $\mathcal{F}_k$ extends the discriminative axis of
the gain $\delta_k$ from question relevance to fact grounding.

\paragraph{Teacher-weight construction and non-decaying gating strength.}
Substituting this Event privilege into the multiplicative basis of \S3.1, we adapt
RLSD's sign-modulated positive weight \citep{yang2026self}, the stop-gradient
teacher-student probability ratio modulated by the advantage sign,
\begin{equation}w_t=\left(\frac{P_T(y_t\mid y_{<t})}{P_S(y_t\mid y_{<t})}\right)^{\operatorname{sign}(A_R)},
\end{equation}where $w_t>1$ amplifies the credit of tokens the Event-privileged teacher favors and
$w_t<1$ suppresses the rest, always in the direction set by $A_R$. Realizing the
memory-level gate $\rho_k$ of \S3.1 at token resolution, we combine the clipped
weight with the identity weight at gating strength $\lambda$,
\begin{equation}\hat A_t=A_R\cdot\rho_t,\qquad \rho_t=(1-\lambda)+\lambda\,\operatorname{clip}(w_t,\,1-\epsilon_w,\,1+\epsilon_w).
\end{equation}Unlike RLSD, which decays $\lambda$ to transition back to standard GRPO, streaming
memory decisions face the invisible-question gap throughout training, so we keep
$\lambda$ constant at $\lambda_0$ and never fade the teacher gate. The clipped
surrogate objective of this branch is thus
\begin{equation}\mathcal{L}_{\text{teacher}}(\hat A)=-\,\mathbb{E}\left[\frac{1}{G}\sum_{i=1}^{G}\frac{1}{|y^{(i)}|}\sum_{t=1}^{|y^{(i)}|}\min\Big(w_t A_R^{(i)},\ \hat A_t^{(i)}\Big)\right].
\end{equation}
\begin{table}[t]
\caption{Results on the real-time subtasks of OVO-Bench and StreamingBench.
Best overall results are in bold and the best results among training-free methods
are underlined.}
\label{tab:perc}
\begin{center}
\scriptsize
\setlength{\tabcolsep}{1.6pt}
\begin{tabular}{p{2.6cm}c|ccccccccccc|ccccccc}
\toprule
\multirow{2}{*}[-0.6ex]{\textbf{Method}} & \multirow{2}{*}[-0.6ex]{\textbf{Size}} & \multicolumn{11}{c|}{\textbf{StreamingBench}} & \multicolumn{7}{c}{\textbf{OVO Real-Time}} \\
\cmidrule(lr){3-13} \cmidrule(lr){14-20}
 & & \textbf{OP} & \textbf{CR} & \textbf{CS} & \textbf{ATP} & \textbf{EU} & \textbf{TR} & \textbf{PR} & \textbf{SU} & \textbf{ACP} & \textbf{CT} & \textbf{Avg.} & \textbf{OCR} & \textbf{ACR} & \textbf{ATR} & \textbf{STU} & \textbf{FPD} & \textbf{OJR} & \textbf{Avg.} \\
\midrule
\multicolumn{20}{l}{\textit{Open-source Offline models}} \\
\midrule
LLaVA-Video~\venue{TMLR'25} & 7B & -- & -- & -- & -- & -- & -- & -- & -- & -- & -- & -- & 69.1 & 58.7 & 68.8 & 49.4 & 74.3 & 59.8 & 63.4 \\
LLaVA-OV~\venue{TMLR'25} & 7B & 80.4 & 74.2 & 76.0 & 80.7 & 72.7 & 71.7 & 67.6 & 65.5 & 65.7 & 45.1 & 71.1 & 66.4 & 57.8 & 73.3 & 53.4 & 71.3 & 62.0 & 64.0 \\
LongVU~\venue{ICML'25} & 7B & -- & -- & -- & -- & -- & -- & -- & -- & -- & -- & -- & 53.7 & 53.2 & 62.9 & 47.8 & 68.3 & 59.8 & 57.6 \\
LongVA~\venue{TMLR'25} & 7B & 70.0 & 63.3 & 61.2 & 70.9 & 62.7 & 59.5 & 61.1 & 53.7 & 54.7 & 34.7 & 60.0 & -- & -- & -- & -- & -- & -- & -- \\
Qwen2.5-VL~\venue{arXiv'25} & 7B & 73.0 & 74.2 & 85.8 & 82.4 & 76.9 & 80.4 & 82.4 & 69.5 & 65.2 & 48.2 & 74.1 & 84.6 & 65.1 & 65.5 & 54.5 & 72.3 & 63.0 & 67.5 \\
Qwen3-VL~\venue{arXiv'25} & 8B & 71.7 & 74.2 & \textbf{89.6} & 84.6 & 68.1 & 85.1 & 83.3 & \textbf{75.6} & 72.0 & 41.5 & 75.7 & 89.3 & \textbf{73.4} & 74.1 & \textbf{64.0} & 73.3 & 64.7 & \underline{73.1} \\
\midrule
\multicolumn{20}{l}{\textit{Open-source Online methods}} \\
\midrule
Dispider~\venue{CVPR'25} & 8B & 74.9 & 75.5 & 74.1 & 73.1 & 74.4 & 59.9 & 76.1 & 62.9 & 62.2 & 45.8 & 67.6 & 57.7 & 49.5 & 62.1 & 44.9 & 61.4 & 51.6 & 54.5 \\
TimeChat-Online~\venue{MM'25} & 7B & 80.2 & 82.0 & 79.5 & 83.3 & 76.1 & 78.5 & 78.7 & 64.6 & 69.6 & \textbf{58.0} & 75.4 & 75.2 & 46.8 & 70.7 & 47.8 & 69.3 & 61.4 & 61.9 \\
FluxMem~\venue{CVPR'26} & 7B & 80.2 & 81.1 & 81.4 & 85.3 & 78.0 & 83.8 & 80.6 & 65.9 & 69.6 & 52.1 & 76.4 & 81.2 & 59.6 & 70.7 & 53.4 & 75.2 & 63.0 & 67.2 \\
StreamForest~\venue{NeurIPS'25} & 7B & 83.1 & \underline{82.8} & 82.7 & 84.3 & 77.5 & 78.2 & 76.9 & 69.1 & \textbf{75.6} & 54.4 & 77.3 & 68.5 & 53.2 & 71.6 & 47.8 & 65.4 & 60.9 & 61.2 \\
Streamo~\venue{CVPR'26} & 7B & -- & -- & -- & -- & -- & -- & -- & -- & -- & -- & -- & 77.2 & 66.1 & \underline{76.7} & 45.5 & 66.3 & \textbf{72.8} & 67.4 \\
VST~\venue{ECCV'26} & 7B & \textbf{85.4} & 82.0 & 86.4 & \textbf{89.1} & 74.2 & \textbf{87.2} & 82.4 & 73.1 & \underline{73.9} & 47.3 & \underline{79.5} & 80.5 & 55.1 & 72.4 & 55.1 & 76.2 & 64.1 & 67.2 \\
\midrule
\multicolumn{20}{l}{\textit{OPSD-based methods}} \\
\midrule
Qwen2.5-VL + SFT & 7B & 73.8 & 75.0 & 85.8 & 82.6 & 73.1 & \underline{86.0} & 75.9 & 72.4 & 68.8 & 49.7 & 75.4 & 81.2 & 70.6 & 67.2 & 52.8 & 77.2 & 63.0 & 68.7 \\
\quad + OPSD~\venue{arXiv'26} &7B & 74.7 & 71.9 & 82.3 & 85.2 & 71.2 & 81.3 & 73.2 & 74.0 & 70.5 & 52.3 & 74.8 & 84.6 & 68.8 & 68.1 & 52.8 & 74.3 & 66.3 & 69.2 \\\quad + RLSD~\venue{arXiv'26} &7B & 76.3 & 76.6 & 86.1 & 84.6 & 75.6 & 82.2 & 73.2 & 72.4 & 72.5 & 51.8 & 76.7 & 83.2 & 64.2 & 69.0 & 53.4 & 76.2 & 65.2 & 68.5 \\
\rowcolor{oursblue}
\quad \textbf{+ EGSD (ours)} & 7B & 79.3 & 80.5 & 88.3 & 81.6 & \textbf{84.4} & 83.2 & \underline{84.3} & 72.0 & 73.7 & \underline{54.9} & 78.4 & 84.6 & 67.9 & 71.6 & 53.9 & \underline{78.2} & 63.6 & 70.0 \\
\cmidrule(lr){1-20}
Qwen3-VL + SFT & 8B & 74.9 & 74.2 & 85.5 & 86.6 & 76.3 & \underline{86.0} & 78.7 & \underline{75.2} & 72.0 & 48.7 & 76.9 & 89.3 & 67.0 & 75.9 & 57.9 & \underline{78.2} & 67.4 & 72.6 \\
\quad + OPSD~\venue{arXiv'26} &8B & 75.2 & 78.9 & 86.4 & 80.3 & 80.0 & 80.7 & 79.6 & 72.4 & 72.2 & 50.3 & 76.0 & \underline{90.6} & 64.2 & 73.3 & 58.4 & 77.2 & 67.4 & 71.9 \\
\quad + RLSD~\venue{arXiv'26} &8B & 82.0 & \textbf{84.4} & 84.5 & 86.2 & 77.5 & 81.6 & 82.4 & \underline{73.6} & 73.1 & 51.8 & 78.2 & 86.6 & 68.8 & \textbf{77.6} & 59.0 & \textbf{80.2} & 67.9 & 73.4 \\
\rowcolor{oursblue}
\quad \textbf{+ EGSD (ours)} & 8B & \underline{83.7} & 79.7 & \underline{89.3} & \underline{87.5} & \underline{81.9} & 83.5 & \textbf{85.2} & 74.0 & 72.8 & 53.9 & \textbf{79.8} & \textbf{91.3} & \underline{71.6} & 72.4 & \underline{60.1} & 76.2 & \underline{69.0} & \textbf{73.4} \\
\bottomrule
\end{tabular}
\end{center}
\end{table}

\subsubsection{Event-Based Entity-Coverage Reward}

The privileged teacher of \S3.2.2 only re-weights tokens the student has already
written, and under token-mean averaging it is invariant to how many reference
entities a memory covers, so it cannot prefer a higher-coverage rollout over a
lower-coverage one. To supply this missing preference we introduce an Event-based entity-coverage reward that ranks rollouts directly by how many effective entities they ground. For the
$k$-th clip, the student memory holds $N_k^{S}$ entities and the Event fact set
$\mathcal{F}_k$ supplies $N_k^{E}$ reference entities; matching the two by one-to-one
semantic correspondence yields $N_k^{+}$
effective entities the student has grounded.
We then define a grounding rate $g_k$ and a coverage rate $\mathrm{cov}_k$,
\begin{equation}g_k = \frac{N_k^{+}}{N_k^{S}},\qquad \mathrm{cov}_k = \frac{N_k^{+}}{N_k^{E}}.
\end{equation}We combine them into a coverage-biased harmonic signal $r_\text{mem}(k)$,
\begin{equation}r_\text{mem}(k) = \frac{(1+\beta_e^2)\,g_k\,\mathrm{cov}_k}{\beta_e^2\,g_k + \mathrm{cov}_k},\qquad \beta_e>1,
\end{equation}where $\beta_e=1.5$ weights coverage above grounding, rewarding entity recall while
still penalizing ungrounded expansion. We center $r_\text{mem}$ within the group at
each clip level to obtain $A_\text{mem}$, added to the loss as an independent
auxiliary policy-gradient signal,
\begin{equation}\mathcal{L} = \mathcal{L}_\text{teacher}(\hat A) + \beta_w\cdot\mathcal{L}_\text{coverage}(A_\text{mem}).
\end{equation}Unlike teacher re-weighting, this signal leaves the sign of $\hat A$ untouched and
ranks rollouts by how many effective entities they ground, so it supplies the
coverage preference the token-mean teacher lacks, with independent
hyperparameters $\beta_e$ balancing grounding against coverage and $\beta_w$
scaling the auxiliary reward against the teacher re-weighting loss. Together, \S3.2.2 and
\S3.2.3 shape streaming memory from two directions, the
privileged teacher enforcing truthfulness and the coverage reward enforcing
sufficiency. Verbosity gaming gains nothing here: each extra entity must still
survive the teacher's gate, so hallucinated additions are down-weighted and the
reward accrues to grounded diversity rather than length.

\section{Experiments}

\begin{table}[t]
\caption{Memory-dependent results on the online video understanding benchmarks
OVO-Bench Backward and Forward, as well as on the offline benchmarks VideoMME
(w/o sub.), LongVideoBench (LVB), and VideoHolmes (VH).}
\label{tab:mem}
\begin{center}
\scriptsize
\setlength{\tabcolsep}{4.4pt}
\begin{tabular}{p{2.4cm}c|ccc|c|ccc|c|cc|c|c}
\toprule
\multirow{3}{*}[-1.2ex]{\textbf{Method}} & \multirow{3}{*}[-1.2ex]{\textbf{Size}} & \multicolumn{8}{c|}{\textbf{Online Video}} & \multicolumn{4}{c}{\textbf{Offline Video}} \\
\cmidrule(lr){3-10} \cmidrule(lr){11-14}
 & & \multicolumn{4}{c|}{\textbf{OVO-Bench Backward}} & \multicolumn{4}{c|}{\textbf{OVO-Bench Forward}} & \multicolumn{2}{c|}{\textbf{VideoMME}} & \multirow{2}{*}[-0.6ex]{\textbf{LVB}} & \multirow{2}{*}[-0.6ex]{\textbf{VH}} \\
\cmidrule(lr){3-6} \cmidrule(lr){7-10} \cmidrule(lr){11-12}
 & & \textbf{EPM} & \textbf{ASI} & \textbf{HLD} & \textbf{Avg.} & \textbf{REC} & \textbf{SSR} & \textbf{CRR} & \textbf{Avg.} & \textbf{Long} & \textbf{Overall} & & \\
\midrule
\multicolumn{14}{l}{\textit{Open-source Offline models}} \\
\midrule
LLaVA-Video~\venue{TMLR'25} & 7B & 56.2 & 57.4 & 7.5 & 40.4 & 34.1 & 70.0 & 60.4 & 54.8 & -- & 63.3 & \textbf{61.3} & -- \\
LLaVA-OV~\venue{TMLR'25} & 7B & 54.2 & 55.4 & 21.5 & 43.7 & 25.6 & 67.1 & 58.8 & 50.5 & -- & 58.2 & 56.3 & -- \\
LongVA~\venue{TMLR'25} & 7B & -- & -- & -- & -- & -- & -- & -- & -- & 47.6 & 54.3 & 56.3 & -- \\
Video-R1~\venue{NeurIPS'25} & 7B & -- & -- & -- & -- & -- & -- & -- & -- & -- & 61.4 & -- & 36.5 \\
LongVILA-R1~\venue{NeurIPS'25} & 7B & -- & -- & -- & -- & -- & -- & -- & -- & 55.2 & 65.1 & 58.0 & -- \\
REVISOR~\venue{CVPR'26} & 7B & -- & -- & -- & -- & -- & -- & -- & -- & \underline{56.2} & \underline{65.7} & 57.5 & -- \\
Qwen2.5-VL~\venue{arXiv'25} & 7B & 50.5 & 67.6 & \underline{42.5} & 53.5 & 29.7 & 50.6 & 48.3 & 42.9 & 46.7 & 54.7 & 56.8 & 33.6 \\
Qwen3-VL~\venue{arXiv'25} & 8B & 55.2 & \textbf{73.0} & 22.6 & 50.3 & 28.4 & 68.4 & 53.3 & 50.0 & 53.8 & 62.6 & 59.6 & 35.9 \\
\midrule
\multicolumn{14}{l}{\textit{Open-source Online methods}} \\
\midrule
Dispider~\venue{CVPR'25} & 8B & 48.5 & 55.4 & 4.3 & 36.1 & 18.1 & 37.4 & 48.8 & 34.7 & -- & 57.2 & -- & -- \\
TimeChat-Online~\venue{MM'25} & 7B & 55.9 & 59.5 & 9.7 & 41.7 & 31.6 & 38.5 & 40.0 & 36.7 & 48.4 & 62.4 & 55.4 & -- \\
StreamForest~\venue{NeurIPS'25} & 7B & 58.9 & 64.9 & 32.3 & 52.0 & 32.8 & 70.6 & 57.1 & 52.5 & -- & 61.4 & -- & -- \\
InfiniPot-V~\venue{NeurIPS'25} & 7B & -- & -- & -- & 47.6 & -- & -- & -- & 47.9 & 53.4 & 59.3 & 56.5 & -- \\
Streamo~\venue{CVPR'26} & 7B & 55.6 & 58.1 & 33.9 & 49.2 & 30.8 & 57.6 & \textbf{82.5} & \textbf{57.0} & -- & -- & -- & -- \\
VST~\venue{ECCV'26} & 7B & 56.9 & 64.9 & \textbf{48.4} & \textbf{56.7} & 33.0 & 66.9 & 62.1 & 54.0 & 55.3 & 64.9 & 58.0 & 41.9 \\
\midrule
\multicolumn{14}{l}{\textit{OPSD-based methods}} \\
\midrule
Qwen2.5-VL + SFT & 7B & 53.2 & 66.9 & 26.9 & 49.0 & 33.1 & 49.9 & 57.9 & 47.0 & 49.7 & 58.2 & 58.6 & 36.7 \\
\quad + OPSD~\venue{arXiv'26} &7B & 53.5 & 61.5 & 23.1 & 46.0 & 24.5 & 62.5 & 56.3 & 47.8 & 52.0 & 60.8 & 57.8 & 43.5 \\
\quad + RLSD~\venue{arXiv'26} &7B & 54.2 & 64.9 & 30.1 & 49.7 & 29.8 & 63.0 & 57.1 & 50.0 & 53.0 & 61.5 & 59.6 & 43.8 \\
\rowcolor{oursblue}
\quad \textbf{+ EGSD (ours)} & 7B & 60.9 & \underline{68.9} & 36.0 & 55.3 & \underline{34.4} & 67.9 & 61.3 & 54.5 & 56.1 & 65.2 & 60.9 & 44.5 \\
\cmidrule(lr){1-14}
Qwen3-VL + SFT & 8B & \underline{61.3} & 64.9 & 32.3 & 52.8 & 24.8 & \underline{70.8} & 64.6 & 53.4 & 54.6 & 63.4 & 60.4 & 43.1 \\
\quad + OPSD~\venue{arXiv'26} &8B & 59.6 & 62.8 & 28.0 & 50.1 & 23.8 & 70.6 & \underline{65.4} & 53.3 & 54.6 & 63.6 & 59.7 & \underline{44.8} \\
\quad + RLSD~\venue{arXiv'26} &8B & 58.6 & 61.5 & 32.3 & 50.8 & 30.7 & \textbf{72.7} & 63.8 & 55.7 & 53.8 & 62.6 & 60.4 & 44.3 \\
\rowcolor{oursblue}
\quad \textbf{+ EGSD (ours)} & 8B & \textbf{62.6} & 66.2 & 39.8 & \underline{56.2} & \textbf{35.2} & \textbf{72.7} & 62.1 & \underline{56.7} & \textbf{57.1} & \textbf{66.5} & \underline{61.1} & \textbf{46.2} \\
\bottomrule
\end{tabular}
\end{center}
\end{table}

\subsection{Implementation Details}

Experiments use \textbf{Qwen2.5-VL-7B-Instruct} \citep{bai2025qwen25vltechnicalreport} and
\textbf{Qwen3-VL-8B-Instruct} \citep{bai2025qwen3} as backbones and sample frames
uniformly at \textbf{1 fps}. Since our memory is text, we first SFT each
backbone on the streaming-memory corpus open-sourced by VST \citep{guan2026video}
to keep the intermediate memory concise and high-quality; all
subsequent RL runs start from this same base checkpoint. Under the streaming
protocol, a split is triggered and a memory write performed every 5000 visual
tokens. All methods are evaluated under this same streaming protocol. During
testing we cap each inference step, including
streaming-think and the final answer, at 8{,}192 video tokens and limit the
thinking count to 4 for efficient evaluation. Low latency is an inherent advantage
of this streaming paradigm rather than an optimization target of our Events; we
report it in Appendix~\ref{app:latency}.

\subsection{Benchmarks and Baselines}

\paragraph{Benchmarks.} We evaluate comprehensively on five video-understanding
benchmarks. Among them, \textbf{StreamingBench} \citep{lin2026streamingbench}
and \textbf{OVO-Bench} \citep{niu2025ovo} are used for online video understanding,
examining online reasoning ability and temporal perception;
\textbf{Video-MME} \citep{fu2025video} is a comprehensive offline benchmark
covering multiple domains and durations; \textbf{LongVideoBench (LVB)}
\citep{wu2024longvideobench} targets long-video understanding ability; and
\textbf{VideoHolmes (VH)} \citep{cheng2025video} focuses on logical reasoning over
video content.

\textbf{Baselines.} For \emph{OPSD-based methods}, we take SFT as the base and
train +\,OPSD \citep{zhao2026self}, +\,RLSD \citep{yang2026self}, and
+\,EGSD each independently on top of the same SFT checkpoint, and also report the
bare backbone Qwen3-VL-8B, evaluating all rows under one
script. \emph{Open-source online}
baselines are Dispider \citep{qian2025dispider}, TimeChat-Online
\citep{yao2025timechat}, StreamForest \citep{zeng2026streamforest}, Streamo
\citep{xia2026streaming}, VST
\citep{guan2026video}, and the training-free FluxMem \citep{xie2026fluxmem} and
InfiniPot-V \citep{kim2026infinipot}, the latter two both built on Qwen2.5-VL-7B.
We also include open-source offline models such as REVISOR \citep{li2026revisor}.

\subsection{Main Results}

\textbf{Results on Online Video Understanding.}
As shown in Table~\ref{tab:perc}, EGSD achieves strong performance in real-time
evaluation. Since most streaming baselines build on Qwen2.5-VL, we compare on that
backbone against the strongest baseline VST. EGSD does not surpass VST on every
StreamingBench subtask, as expected since its Event optimization targets cross-time
memory, not single-frame perception. It wins on the four subclasses that
require accumulating events across segments, event understanding (EU, +10.2),
temporal counting (CT, +7.6), clip summarization (CS, +1.9), and prospective
reasoning (PR, +1.9) over VST, covering 778 of 2498 questions (31\%). On the
remaining single-frame subclasses it largely holds or improves over the backbone, so
stronger memory does not erode perception, and on the OVO-Bench Real-Time track it
leads VST by 2.8 on average.

\textbf{Results on Memory-Dependent Understanding.}
Table~\ref{tab:mem} evaluates the streaming OVO-Bench Backward and Forward tracks,
whose queries target events already streamed past, together with the offline
VideoMME, LongVideoBench, and VideoHolmes, where our core advantage concentrates.
Comparing on Qwen2.5-VL against the strongest baseline VST, EGSD improves the
Backward subsets EPM by 4.0 and ASI by 4.0. On Forward it
leads by 0.5, and on the offline benchmarks it improves VideoMME-Overall by 0.3,
LongVideoBench by 2.9, and VideoHolmes by 2.6. Against its own SFT start the gains
are far larger, with the Backward average up 6.3, the Forward average up 7.5,
VideoMME-Overall up 7.0, and VideoHolmes up 7.8. Because VideoHolmes is a purely
offline reasoning benchmark, this shows the memory from teacher distillation and the
entity-coverage reward transfers beyond streaming without trading off offline quality.

\textbf{Effect of Model Size and the OPSD Signal.}
The gains persist across backbone scale. On Qwen3-VL-8B, EGSD leads every memory
benchmark on its overall metric, lifting the OVO-Bench Backward / Forward averages
from 52.8 / 53.4 to 56.2 / 56.7, VideoMME-Overall from 63.4 to 66.5,
LongVideoBench from 60.4 to 61.1, and VideoHolmes from 43.1 to 46.2. Comparing the
OPSD and RLSD rungs then isolates our first design choice. Vanilla OPSD often falls
below its SFT start on the memory columns, with Backward dropping 49.0 to 46.0 on
Qwen2.5-VL, because its teacher scores memory with the dual question and GT known
while the student wrote it before any question arrived, misaligning its preference
with optimizing memory for the answer. RLSD instead combines the same signal
multiplicatively with the outcome reward and turns most memory columns positive
again, confirming the multiplicative form is the right way to apply the OPSD signal
under the streaming protocol; EGSD builds on this form and pushes the gains further.


\subsection{Ablation Studies of EGSD Components}

\begin{wraptable}{r}{0.58\textwidth}
\vspace{-\baselineskip}
\caption{Ablation starting from Vanilla RLSD, enabling components one by one. Backbone Qwen3-VL-8B.}
\label{tab:ablation}
\centering
\scriptsize
\setlength{\tabcolsep}{2.3pt}
\begin{tabular}{p{3.05cm}ccccc}
\toprule
\multirow{2}{*}[-0.6ex]{\textbf{Configuration}} & \multicolumn{3}{c}{\textbf{OVO-Bench}} & \multirow{2}{*}[-0.6ex]{\textbf{Video-MME}} & \multirow{2}{*}[-0.6ex]{\textbf{VideoHolmes}} \\
\cmidrule(lr){2-4}
 & \textbf{BT} & \textbf{FW} & \textbf{Avg.} & & \\
\midrule
Vanilla RLSD & 50.8 & 55.7 & 53.3 & 62.6 & 44.3 \\
\quad + remove decay rate & 52.1 & 55.3 & 53.7 & 63.8 & 44.9 \\
\quad + privileged Event & \underline{55.3} & \underline{56.1} & \underline{55.7} & \underline{65.6} & \textbf{46.5} \\
\midrule
\quad \textbf{+ entity-coverage (EGSD)} & \textbf{56.2} & \textbf{56.7} & \textbf{56.5} & \textbf{66.5} & \underline{46.2} \\
\bottomrule
\end{tabular}
\vspace{-\baselineskip}
\end{wraptable}

As shown in Table~\ref{tab:ablation}, starting from Vanilla RLSD, removing the
decay-rate term lifts OVO-Bench, Video-MME, and VideoHolmes by consistent margins,
showing the teacher should supervise the whole streaming process. Adding the Event
privileged teacher gives the largest single jump, raising OVO-Backward,
OVO-Forward, and their average. The complete EGSD then adds the entity-coverage
reward and reaches the ladder-highest OVO-Bench average and Video-MME with
VideoHolmes level. The three components are complementary and synergistic.

\subsection{Memory Quality Analysis}

\begin{figure}[t]
\centering
\includegraphics[width=\linewidth]{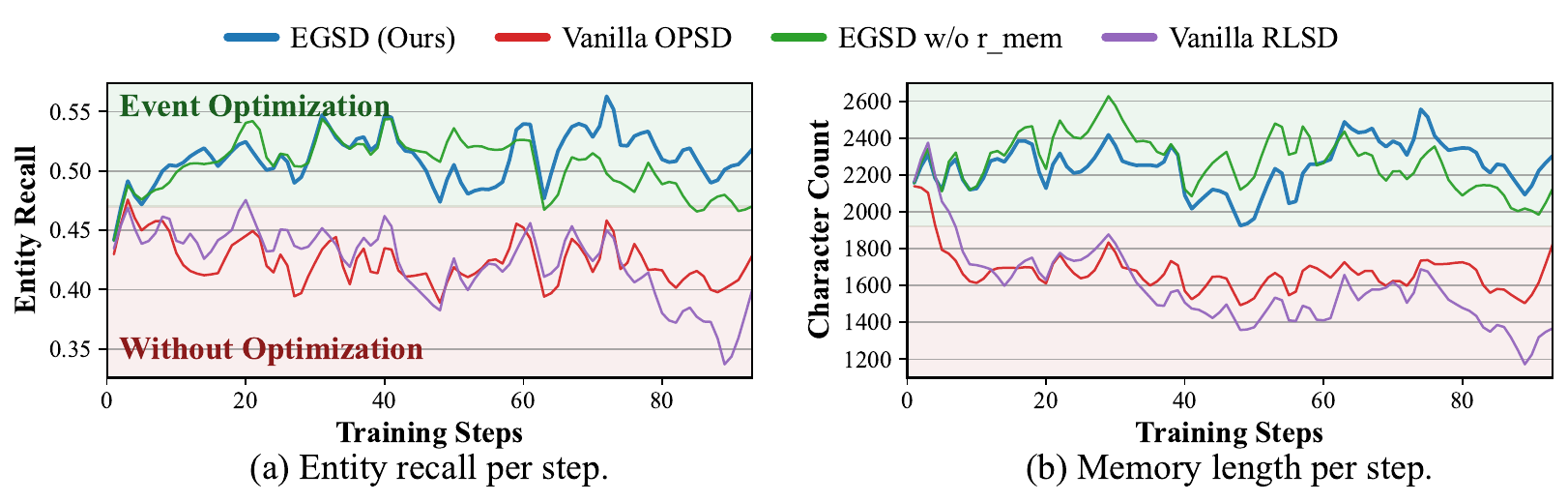}
\caption{Per-step entity recall (a) and memory character count (b) for four methods
sharing the same SFT start.}
\label{fig:memory_quality}
\end{figure}

This section characterizes the quality of the memory itself. To separate
memory getting longer from memory getting better, we record per step two intrinsic
indicators, entity recall, the fraction of the reference Event entities of \S3.2.3
that the student trajectory grounds, and the average memory character
count for length overhead. The four compared methods are Vanilla OPSD, Vanilla
RLSD, EGSD, and EGSD without $r_\text{mem}$, the ablation of \S4.4.

\begin{figure}[t]
\begin{center}
\includegraphics[width=\textwidth]{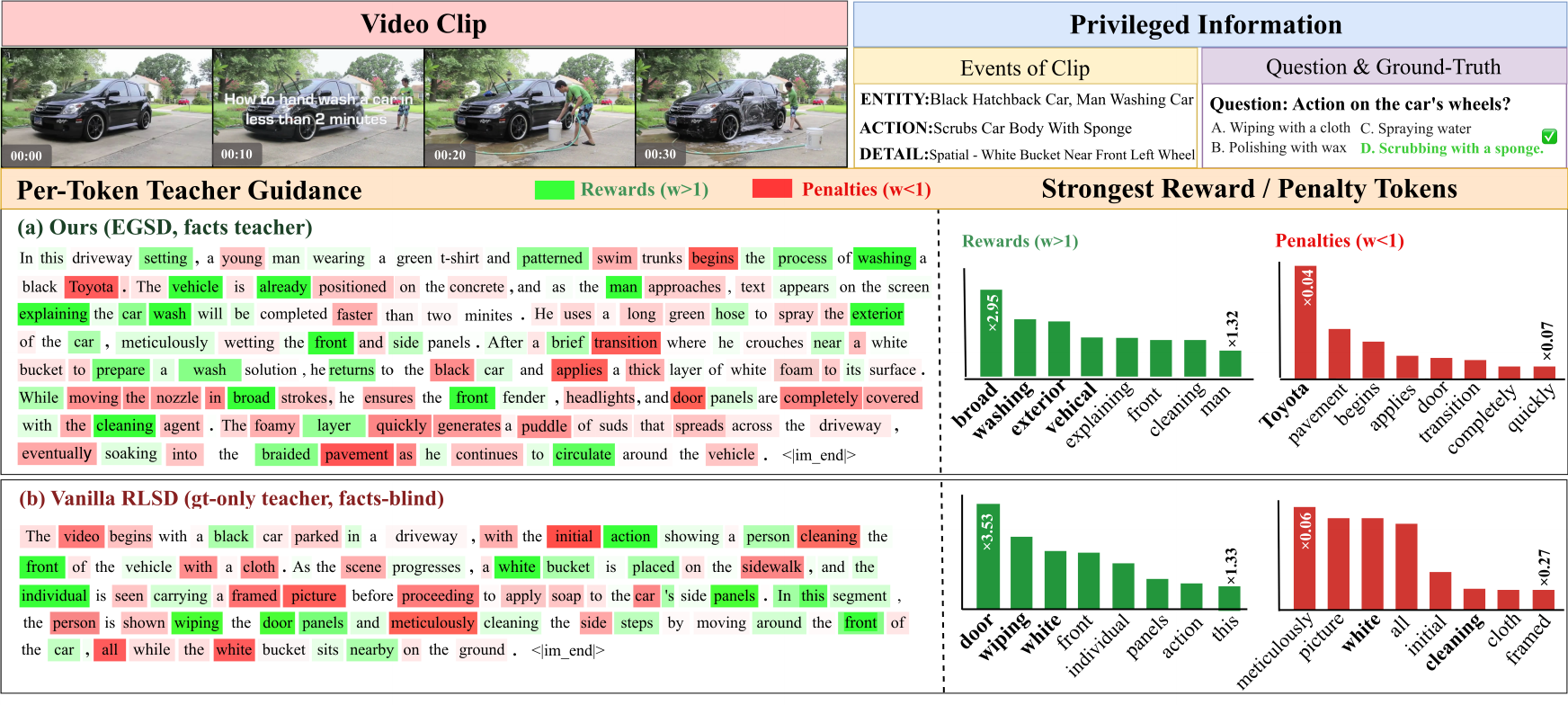}
\end{center}
\caption{Case study of per-token teacher guidance on the same clip. (a) EGSD teacher with Event facts; (b) Vanilla RLSD teacher with only the question and ground-truth answer.}
\vspace{0\baselineskip}
\label{fig:carwash}
\end{figure}

\subsubsection{Recall and Memory-Length Analysis}

Figure~\ref{fig:memory_quality} shows the per-step curves of entity recall and
memory character count. On recall, the Event-privileged teacher and coverage
reward $r_\text{mem}$ jointly fill the teacher's blind spot, so EGSD climbs
to $0.518$ ($+17.4\%$ over the SFT start) while Vanilla RLSD instead drops to $0.400$
($-9.4\%$), a $29.5\%$ gap
at convergence. This gain does not come from writing more: EGSD's memory grows only
to $2303$ characters, $+6.8\%$ over the shared SFT start, with no runaway growth.
Removing $r_\text{mem}$ confirms its role, recall falls from $0.518$ to $0.470$
($-9.3\%$) and length from $2303$ to $2124$ characters, so $r_\text{mem}$ raises
recall while adding only marginal memory.

\subsubsection{Tokens Rewarded Analyse}

\begin{wraptable}{r}{0.5\textwidth}
\vspace{-4\baselineskip}
\caption{Reward composition: visual-entity ratio of rewarded tokens across training steps.}
\label{tab:visual-entity}
\centering
\scriptsize
\setlength{\tabcolsep}{4pt}
\begin{tabular}{lccccc}
\toprule
\textbf{Step} & \textbf{10} & \textbf{30} & \textbf{50} & \textbf{70} & \textbf{90} \\
\midrule
Vanilla RLSD & 0.204 & 0.280 & 0.237 & 0.288 & 0.300 \\
\textbf{EGSD (ours)} & \textbf{0.366} & \textbf{0.325} & \textbf{0.306} & \textbf{0.380} & \textbf{0.356} \\
\bottomrule
\end{tabular}
\vspace{-2\baselineskip}
\end{wraptable}

This section analyzes the tokens rewarded by the teacher, using a macro visual-entity
ratio across training together with a per-token guidance case study.

\textbf{Visual-entity ratio of rewarded tokens.} We define visual entities as
the union of \texttt{ENTITY} words registered for the clip in the fact cache and
measure the fraction of teacher-rewarded content words landing on this union
(Table~\ref{tab:visual-entity}). EGSD's visual-entity ratio is clearly higher than
Vanilla RLSD throughout training, with the gap stably maintained. We never set this
ratio as a training target; it emerges after re-weighting the teacher with Event
facts, which pushes gradient weight toward entities truly present in the scene and
shapes memory-writing habits early.

\textbf{Case study.} We place the Vanilla RLSD and EGSD teachers on the
same clip, differing only in privileged information. On
a clip of a man hand-washing a black car, asked what action the person performs on
the wheels, the registered facts cover the black car, the man in green, a bucket, a
hose, and wiping the body with a sponge. As shown in
Figure~\ref{fig:carwash}, the two teachers reward very different tokens. EGSD
concentrates reward on the valid visual entities and actions truly present in the
scene, giving high weight to the registered washing action
(\texttt{broad}$\times$2.95, \texttt{wash}$\times$2.05, \texttt{exterior}$\times$2.00)
and to grounded objects, so it encourages the student to record more effective
entities, while its deepest penalty $\times$0.04 falls only on the brand
hallucination \texttt{Toyota}. Vanilla RLSD instead penalizes salient visual
entities only weakly tied to the question, scoring \texttt{car}$\times$0.806 and
\texttt{cleaning}$\times$0.713 where EGSD rewards the same tokens at
\texttt{cleaning}$\times$1.569 and \texttt{car}$\times$1.249. Its guidance is even
unstable within one clip, rewarding \texttt{white} in the white bucket at
$\times$1.996 on first mention but $\times$0.079 on the second, since a
question-and-GT privilege cannot judge whether an entity should be recorded for
being salient in the scene or only for being relevant to the question.
Lacking effective facts, the vanilla teacher penalizes salient
entities irrelevant to the question and collapses memory, whereas
the Event fact teacher grounds the gradient in real facts and lifts entity recall.

\section{Conclusion}

This paper introduces EGSD, an on-policy self-distillation method that optimizes
streaming video memory end-to-end. Characterizing memory as an incremental update over
verifiable Events, EGSD combines multiplicative teacher gating, Event-privileged
re-weighting, and an entity-coverage reward to counter the optimization failure and memory
collapse of vanilla OPSD. Across online and offline benchmarks, EGSD improves both memory
diversity and answering accuracy for streaming video understanding, and we hope it offers
useful insights for future work in this direction.

\subsection*{AI use statement}

In this work, generative AI tools were used solely to polish the writing of parts
of the paper, correcting grammar and improving fluency of author-written text. They
were not used for research ideation, experimental design, implementation, data
analysis, or generating any scientific claims or results. All AI-assisted text was
reviewed by the authors, who take full responsibility for the final content of this
work.

\bibliography{custom}
\bibliographystyle{iclr2027_conference}

\appendix
\section{Inference Latency}
\label{app:latency}

We report two latency notions and, following VST's Table~6, keep them apart.
\emph{TTFT} (Time-To-First-Token) is the delay from posing the question to the
first answer token, and is the operative metric for direct-answer and online
streaming methods. \emph{End-to-end QA latency} (E2E) runs from the end of video
input to the last token of a full answer, and is the only faithful metric for
offline with-CoT methods, whose reasoning chain is produced \emph{after} the
question --- reporting TTFT for them would undercount the true cost by an order of
magnitude. All ``ours'' rows are measured on one harness and GPU batch with
\texttt{failure\_rate}=0; the rest are quoted from the VST paper.

\begin{table}[!ht]
\caption{Inference latency on VideoHolmes (p50), reported for both the
Qwen2.5-VL-7B and Qwen3-VL-8B backbones. Offline with-CoT rows report end-to-end
QA latency (E2E); direct-answer and online rows report TTFT. Rows shaded green are
offline, orange online; numbers not marked ``ours'' are quoted from the VST paper.}
\label{tab:latency}
\centering
\small
\setlength{\tabcolsep}{8pt}
\begin{tabular}{ll c c}
\toprule
\textbf{Type} & \textbf{Method} & \textbf{Metric} & \textbf{Latency (s)} \\
\midrule
\rowcolor{green!12} Offline & Qwen2.5-VL-7B w/CoT & E2E & $5.30$ \\
\rowcolor{green!12} Offline & Video-R1 w/CoT & E2E & $8.80$ \\
\rowcolor{green!12} Offline & Qwen2.5-VL-7B w/CoT (ours repro) & E2E & $5.68$ \\
\rowcolor{green!12} Offline & Qwen3-VL-8B w/CoT (our 8B backbone) & E2E & $5.89$ \\
\rowcolor{green!12} Offline & Qwen2.5-VL-7B direct-answer & TTFT & $0.54$ \\
\midrule
\rowcolor{orange!12} Online & Dispider-7B & TTFT & $1.10$ \\
\rowcolor{orange!12} Online & VST-7B & TTFT & $0.56$ \\
\rowcolor{orange!12} Online & VST-32B & TTFT & $1.40$ \\
\rowcolor{oursblue} Online & \textbf{EGSD-7B (ours)} & TTFT & $0.90$ \\
\rowcolor{oursblue} Online & \textbf{EGSD-8B (ours)} & TTFT & $0.73$ \\
\bottomrule
\end{tabular}
\end{table}

Holding the backbone fixed, the two regimes diverge sharply. On Qwen3-VL-8B, an
offline with-CoT answer needs $5.89$\,s end-to-end, whereas our online streaming
memory returns a first token in $0.73$\,s ($\approx8\times$ faster); on
Qwen2.5-VL-7B the contrast is $5.68$ vs $0.90$\,s ($\approx6\times$). The gain is
structural, not a smaller-model artifact: the streaming think is front-loaded,
completing asynchronously as clips arrive before the question rather than stacking
after it, so our TTFT stays in the same sub-second range as VST-7B ($0.56$\,s) even
with the added memory mechanism. Across seven benchmarks the sample-weighted median
first token is $0.98$\,s (8B), on par with the GT-only variant, so the memory shaped
by $r_{\mathrm{mem}}$ costs no extra latency. Because VST rows come from different
hardware, we compare orders of magnitude and the online--offline gap, not absolute
seconds.

\end{document}